\documentclass[sigconf]{acmart}
  \renewcommand\footnotetextcopyrightpermission[1]{} % removes footnote with conference information in first column
\usepackage{booktabs} % For formal tables
\usepackage{algorithm}  
\usepackage{algorithmic}

\setcopyright{rightsretained}
\begin{document}
\title{Cross-Domain Adversarial Auto-Encoder}
% \titlenote{Produces the permission block, and
% \subtitle{Extended Abstract}
% \subtitlenote{The full version of the author's guide is available as
%   \texttt{acmart.pdf} document}

\author{Haodi Hou}
% \authornote{Dr.~Trovato insisted his name be first.}
\affiliation{%
  \institution{State Key Laboratory for Novel Software Technology, Nanjing University, China}
  \streetaddress{No. 163 Xianlin Avenue}
  \city{Nanjing}
  \state{Jiangsu}
  \postcode{210046}
}
\email{hhd@smail.nju.edu.cn}

\author{Jing Huo}
% \authornote{The secretary disavows any knowledge of this author's actions.}
\affiliation{%
  \institution{State Key Laboratory for Novel Software Technology, Nanjing University, China}
  \streetaddress{No. 163 Xianlin Avenue}
  \city{Nanjing}
  \state{Jiangsu}
  \postcode{210046}
}
\email{huojing@nju.edu.cn}

\author{Yang Gao}
% \authornote{This author is the
%   one who did all the really hard work.}
\affiliation{%
  \institution{State Key Laboratory for Novel Software Technology, Nanjing University, China}
  \streetaddress{No. 163 Xianlin Avenue}
  \city{Nanjing}
  \state{Jiangsu}
  \postcode{210046}
}
\email{gaoy@nju.edu.cn}

% \renewcommand{\shortauthors}{B. Trovato et al.}

\begin{abstract}
  In this paper, we propose the Cross-Domain Adversarial Auto-Encoder (CDAAE) to address the problem of cross-domain image inference, generation and transformation. We make the assumption that images from different domains share the same latent code space for content, while having separate latent code space for style. The proposed framework can map cross-domain data to a latent code vector consisting of a content part and a style part. The latent code vector is matched with a prior distribution so that we can generate meaningful samples from any part of the prior space. Consequently, given a sample of one domain, our framework can generate various samples of the other domain with the same content of the input. This makes the proposed framework different from the current work of cross-domain transformation. Besides, the proposed framework can be trained with both labeled and unlabeled data, which makes it also suitable for domain adaptation. Experimental results on data sets SVHN~\cite{svhn}, MNIST~\cite{mnist} and CASIA~\cite{casia} show the proposed framework achieved visually appealing performance for image generation task. Besides, we also demonstrate the proposed method achieved superior results for domain adaptation. Code of our experiments is available in \url{https://github.com/luckycallor/CDAAE}.
\end{abstract}

  


\maketitle

\section{Introduction}

Appealing results have been achieved in computer vision tasks with data from one single domain. However, in many real applications, it's common that data is from different domains, where samples share the same content with quit different appearance. Such examples include, same characters in different fonts, same objects represented by text and image, etc. There are still many problems when dealing with cross-domain data. In this paper, we mainly consider three problems about cross-domain data: inference, generation/transformation and domain adaptation. A compact framework which is able to do inference, generation/transformation for cross-domain data is firstly presented. Then we further extended the framework and proposed an algorithm to address the problem of domain adaptation.

As one of the basic tasks in machine learning, performing an inference mechanism allows reasoning about data at an abstract level~\cite{ali}. However, most existing works of inference mechanism mainly focus on single domain data. For example, to deal with intractable posterior distributions of directed probabilistic models, Variational Auto-Encoder (VAE)~\cite{vae} trains the auto-encoder by minimizing the reconstruction loss and uses a KL-divergence penalty to impose a prior distribution on the latent code vector. Adversarial Auto-Encoders (AAE)~\cite{aae} use an adversarial training criterion to match the aggregated posterior of the latent code vector with a prior distribution, which allows to impose more complicated distributions. As variants of Generative Adversarial Nets (GANs)~\cite{gan}, Adversarially Learned Inference (ALI)~\cite{ali} or Bidirectional GANs~\cite{BidirectionalGAN} cast the learning of an inference machine and a deep directed generative model in a GAN-like adversarial framework, where the discriminators examine joint (data, latent) pairs. Others~\cite{Semi-DGM,aae,UnLofIFH,bilinear} also explore learning a representation disentangling content and style. As for cross-domain data, UNsupervised Image-to-image Translation (UNIT)~\cite{unI2I} tries to learn representation of data from both domains in a totally shared latent code space. While in practice, cross-domain data of the same class only shares same content, but with quite different styles. Therefore, there is more to explore for inference learning of cross-domain data.

In the field of image generation, GANs can generate vivid novel samples matching the given data set. When data is from one domain, generative models mainly refer to those map a latent code drawn from a prior distribution to samples from the data distribution. But when data is from different domains, the latent code can be either drawn from a prior distribution or gained by encoding data of another domain. The latter is another hot field of research, cross-domain transformation, where many models have achieved appealing results. Style transfer technologies~\cite{styleTransfer-Gatys,styleTransfer-Johnson} are able to migrate the visual style from one image to another, but they mostly ignore the semantic information. Image analogy~\cite{imageAnalogy,IA-1,IA-2,imageAnalogy-ms} can generate analogical images with different visual attributes while maintaining the semantic structure. GAN-based methods~\cite{cycleGAN,dtn,unI2I} are able to achieve cross-domain transformation without paired examples. Though satisfactory results are achieved, these image transformation methods mostly share the same shortcoming that they can only generate one corresponding image with a certain input, as there is no explicitly disentangled inference of content and style in latent code space.

Another problem in the field of cross-domain data is domain adaptation. Since collecting well-annotated dataset is prohibitively expensive, an appealing alternative is to generalize models learned with labeled data from one domain to another domain without labels. The work of Basura et. al.~\cite{domainAdapt-subspace} implements domain adaptation through subspace learning, while some other works~\cite{dann,CoGAN,dtn,dsn,pixelDA} make use of the deep feature learning. As our framework is designed to be trained with both labeled and unlabeled data, there is an innate advantage to do domain adaptation. Therefore, a domain adaptation algorithm based on the proposed framework is also proposed. Experimental results show that state-of-the-art accuracy is achieved on benchmark data sets.

To deal with the above problems, we propose a novel framework called Cross-Domain Adversarial Auto-Encoder (CDAAE). The basic assumption of the proposed framework is that data can be disentangled into content and style. The content part decides the label of the data and the style part influences the appearances or attributes of the data. Then the proposed framework further adopts the assumption that data from different domains have a shared latent code space for content and different latent code spaces for styles. In the encoding process, our framework maps data from two domains into a latent code vector consisting of a content part and a style part. The content part is shared by data from both domains and can be trained with both labeled and unlabeled data, while the style part is separate for different domains and is trained in a totally unsupervised manner. In the decoding process, two generators are trained to map a latent code vector to data of corresponding domains with the corresponding style latent code. The CDAAE can learn an approximate inference of data from two domains and disentangle content and style. In generation or transformation process, the style part in latent code vector can be draw from the prior distribution, which allows CDAAE to generate various samples with a certain input. This makes the proposed framework different from the previous works. Besides, as the proposed framework can be trained with both supervised and unsupervised data, it is also suited for domain adaptation. Experimental results demonstrate the domain adaptation algorithm based on CDAAE achieved state-of-the-art accuracy on benchmark data sets.

\section{Related Works}

GANs have achieved impressive results in image generation, and a series of variants have been proposed to complete different tasks, such as image inference~\cite{ali,BidirectionalGAN}, conditional image generation~\cite{conditionalGAN,infogan,inpainting,text2image}, unsupervised learning~\cite{pixelDA,unsupervised} and cross-domain transformation~\cite{cycleGAN,I2I-conditional,unI2I}. All these variants maintain the idea of using adversarial loss to measure the difference between distributions.

There are mainly two ways to complete image transformation. One is by a pixel-to-pixel way that learns a direct mapping from image to image, such as CycleGAN~\cite{cycleGAN} and work of Phillip et. al.~\cite{I2I-conditional}. The other one is by an encoding-decoding way that explicitly maps images from one domain to latent codes first, and then decodes the latent codes to images of the other domain, such as Domain Transfer Network (DTN)~\cite{dtn} and UNIT~\cite{unI2I}.

The proposed framework is organized in an encoding-decoding manner and is trained with an adversarial training criterion. Different from the previous inference learning works, our framework learns a latent code vector of cross-domain data and disentangles the content and style. In contrast to other image transformation works, our framework can generate various samples instead of a fixed one with a certain input.

\subsection{Auto-Encoders}

Auto-encoders are a series of deep neural networks that consist of an encoder and a decoder (or generator). The encoder maps an image into a latent code vector which follows the prior distribution, and this is usually achieved by minimizing the difference between the learned posterior distribution and the prior distribution. The decoder inversely maps a latent code vector into an image and is trained by minimizing the reconstruction loss. This could be described as the following optimization problem:
\begin{equation}
  \min_{f_e,f_g} E_{\boldsymbol{x} \sim p_{data}}[d_1(p_{f_e(\boldsymbol{x})}, p_{prior})]+E_{\boldsymbol{x} \sim p_{data}}[d_2(\boldsymbol{x},f_g(f_e(\boldsymbol{x})))],
\end{equation}
where $\boldsymbol{x}$ is a sample from data distribution $p_{data}$. The first term is the difference between prior distribution $p_{prior}$ and that of latent code vectors generated by encoder $p_{f_e(\boldsymbol{x})}$, and $d_1$ is a distance function of two distributions (KL-divergence for VAE~\cite{vae} and adversarial loss for AAE~\cite{aae}). $f_e$ denotes the encoder function represented by a multilayer perceptron. The second term measures the reconstruction loss, where $d_2$ is a distance function of two images (such as Euclidean distance), and $f_g$ denotes the generator function represented by another multilayer perceptron.

\subsection{Adversarial Loss}

The raise of GANs~\cite{gan} inspires us to use adversarial loss to measure the difference of two distributions. The main idea of GANs is to improve both the generator and discriminator simultaneously through a two-player minimax game. This could be analogized as following: the generator is a forger which creates forgeries and tries to make them as realistic as possible, while the discriminator is an expert who receives both forgeries and real images and tries to tell them apart. In the competition between the forger and expert, they both improve their abilities. The minimax game between the generator and the discriminator is as follows:
\begin{equation}
  \min_{f_g}\max_{f_d}E_{\boldsymbol{x} \sim p_{data}}[\log f_{d}(\boldsymbol{x})]+E_{\boldsymbol{z} \sim p_{prior}}[\log(1-f_{d}(f_{g}(\boldsymbol{z})))].
\end{equation}

$f_g$ and $f_d$ are differentiable functions of the generator and discriminator represented by two multilayer perceptrons respectively. $\boldsymbol{x}$ is sample from the distribution of real data $p_{data}$, while $\boldsymbol{z}$ is the initial latent code vector drawn from the prior distribution $p_{prior}$. In the maximization process, discriminator's ability to tell real data and data generated by generator is improved, while in the minimization process, the ability of generator to fool the discriminator is improved. Consequently, the discriminator can tell real data from others and the generator can generate indistinguishable examples from real data. In other words, the distribution of the generator is the same as that of the real data. 

\begin{figure*}[h]
  \centering
  \includegraphics{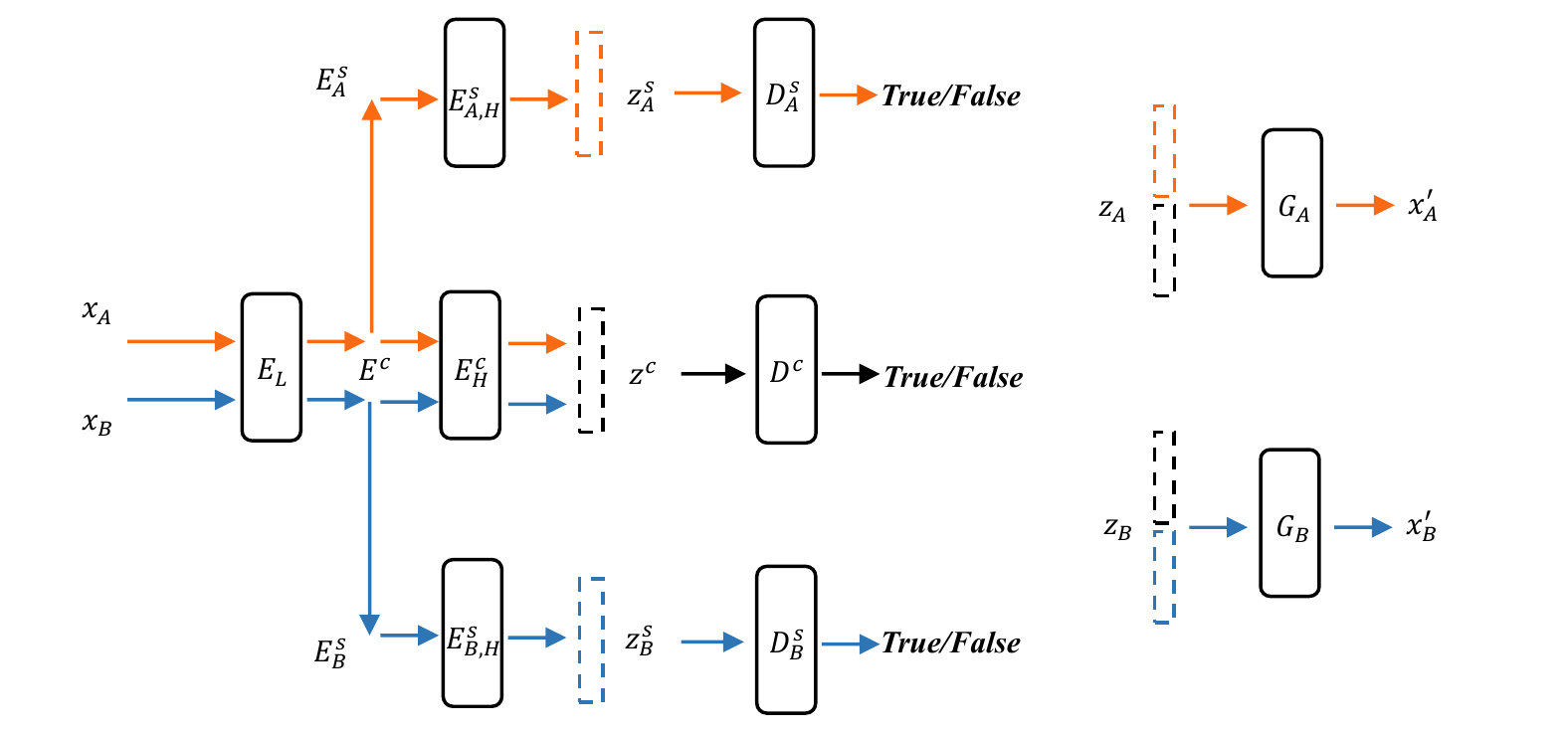}
  \caption{In the framework above, orange lines denote the flow of images and latent code vectors of domain A, while blue ones denote that of domain B. Rounded rectangles denote neural networks and dashed ones denote latent code vectors. $E_L$ and $E_H^c$ are low-level and high-level layers of the content-latent-code encoder. $E_{A,H}^s$ and $E_{B,H}^s$ are high-level layers of the style-latent-code encoders for domain A and B respectively. The content-latent-code encoder $E^c$ consists of $E_L$ and $E_H^c$, and $E_L$ is shared with the style-latent-code encoders $E_A^s$ and $E_B^s$. $D^c$ is the discriminator to tell whether the input is drawn from the prior distribution or generated by the encoder, and plays an adversarial game with the encoder. So do $D_A^s$ and $D_B^s$. $G_A$ and $G_B$ are generators for images of domain A and B respectively with input of content-latent-code concatenated with the corresponding style-latent-code.}
  \label{framework}
\end{figure*}

\section{Cross-Domain Adversarial Auto-Encoder}

Assume that $A$ and $B$ are two image domains which share the same content. In image inference, image examples drawn from marginal distributions $P_A(\boldsymbol{x}_A)$ and $P_B(\boldsymbol{x}_B)$ are given. Denote $\boldsymbol{x}_A$ and $\boldsymbol{x}_B$ two samples from domain $A$ and $B$ respectively. To complete this task, we suppose there are two latent code spaces $Z_A$ and $Z_B$ with prior distributions $Q_A(\boldsymbol{z}_A)$ and $Q_B(\boldsymbol{z}_B)$ for the two domains respectively. $\boldsymbol{z}_A$ and $\boldsymbol{z}_B$ denote samples drawn from the prior distributions. Then the inference task can be formulated as learning the conditional distributions $p_A(\boldsymbol{z}_A|\boldsymbol{x}_A,\theta_A^e)$ and $p_B(\boldsymbol{z}_B|\boldsymbol{x}_B,\theta_B^e)$ with learning target:
\begin{equation}\begin{split}
  \min_{\theta_A^e, \theta_B^e}&d_1(Q_A(\boldsymbol{z}_A),\int_{\boldsymbol{x}_A}P_A(\boldsymbol{x}_A)p_A(\boldsymbol{z}_A|\boldsymbol{x}_A,\theta_A^e))\\+&d_1(Q_B(\boldsymbol{z}_B),\int_{\boldsymbol{x}_B}P_B(\boldsymbol{x}_B)p_B(\boldsymbol{z}_B|\boldsymbol{x}_B,\theta_B^e)),
\end{split}\end{equation} 
where $d_1$ is a measurement of the difference between two distributions. $\theta_A^e$ and $\theta_B^e$ are the parameters of the corresponding conditional distributions.

On the other side, the generation task can be formulated as learning the conditional distributions $q_A(\boldsymbol{x}_A|\boldsymbol{z}_A,\theta_A^g)$ and $q_B(\boldsymbol{x}_B|\boldsymbol{z}_B,\theta_B^g)$ by minimizing the reconstruction loss:
\begin{equation}\begin{split}
  \min_{\theta_A^g,\theta_B^g}&E_{\boldsymbol{x}_A,\boldsymbol{z}_A}[d_2(\boldsymbol{x}_A,p_A(\boldsymbol{z}_A|\boldsymbol{x}_A,\hat{\theta}_A^e)q_A(\boldsymbol{x}_A|\boldsymbol{z}_A,\theta_A^g))]\\+&E_{\boldsymbol{x}_B,\boldsymbol{z}_B}[d_2(\boldsymbol{x}_B,p_B(\boldsymbol{z}_B|\boldsymbol{x}_B,\hat{\theta}_B^e)q_B(\boldsymbol{x}_B|\boldsymbol{z}_B,\theta_B^g))],
\end{split}\end{equation}
where $d_2$ is a distance function of original image and reconstructed image. $\hat{\theta}_A^e$ and $\hat{\theta}_B^e$ are the optimal parameters for the inference task. $\theta_A^g$ and $\theta_B^g$ are parameters to be learned for the corresponding conditional distributions, and their optima are denoted as $\hat{\theta}_A^g$ and $\hat{\theta}_B^g$ respectively.

Up to now, we have formulated the tasks of inference and generation. As for the transformation, we need to build connections between the images or latent codes of two domains. Since an infinite set of possible joint distributions can yield the given marginal distributions of either images or latent code, additional assumptions are needed. As the two domains share the same content, we can assume that the marginal distributions of latent code have a shared content-latent-code space. Therefore the latent code consists of two parts: the shared content-latent-code and the unshared style-latent-code. The latent code and the corresponding distribution become as follows:
\begin{equation}
  \boldsymbol{z}_A=[\boldsymbol{z}^c,\boldsymbol{z}_A^s],
\end{equation}
\begin{equation}
  \boldsymbol{z}_B=[\boldsymbol{z}^c,\boldsymbol{z}_B^s],
\end{equation}
\begin{equation}
  Q_A(\boldsymbol{z}_A) = Q^c(\boldsymbol{z}^c)Q_A^s(\boldsymbol{z}_A^s),
\end{equation}
\begin{equation}
  Q_B(\boldsymbol{z}_B) = Q^c(\boldsymbol{z}^c)Q_B^s(\boldsymbol{z}_B^s),
\end{equation}
where $\boldsymbol{z}^c$ is the shared content latent code and $Q^c(\boldsymbol{z}^c)$ is its prior distribution. The superscript $c$ denotes content. $\boldsymbol{z}_A^s$, $\boldsymbol{z}_B^s$ are the style latent code for domain A, B respectively, and $Q_A^s(\boldsymbol{z}_A^s)$, $Q_B^s(\boldsymbol{z}_B^s)$ are their corresponding prior distributions. The superscript $s$ denotes style. The conditional distributions for inference task correspondingly become:
\begin{equation}
  p_A(\boldsymbol{z}_A|\boldsymbol{x}_A,\theta_A^e)=p^c(\boldsymbol{z}^c|\boldsymbol{x}_A,\theta^{e,c})p_A^s(\boldsymbol{z}_A^s|\boldsymbol{x}_A,\theta^{e,s}_A),
\end{equation}
\begin{equation}
  p_B(\boldsymbol{z}_B|\boldsymbol{x}_B,\theta_B^e)=p^c(\boldsymbol{z}^c|\boldsymbol{x}_B,\theta^{e,c})p_B^s(\boldsymbol{z}_B^s|\boldsymbol{x}_B,\theta^{e,s}_B),
\end{equation}
where $p^c$ is the conditional distribution for the content latent code, and $p_A^s$, $p_B^s$ are the conditional distribution for the style latent code of domain $A$, $B$. $\theta^{e,c}$, $\theta^{e,s}_A$, $\theta^{e,s}_B$ are their parameters to learn. $\theta_{A}^{e}$ is composed of $\theta^{e,c}$ and $\theta_{A}^{e,s}$. Similarly, $\theta_{B}^{e}$ is composed of $\theta^{e,c}$ and $\theta_B^{e,s}$.   

To achieve cross-domain transformation, we can firstly calculate the content-latent-code of the input image, and then concatenate it with a style-latent-code drawn from the prior style-latent-code distribution of the other domain, and feed the result into the generator of the other domain to get an output image. With various style-latent-code instances drawn from the prior distribution, we can generate various images of the other domain just using one certain input image.

\subsection{Framework}

Based on the assumption and formulation above, our framework is illustrated in Figure~\ref{framework}. The framework mainly consists of two parts: one is the encoders which are trained by playing an adversarial game with the corresponding discriminators and the other is the generators, which are trained by minimizing the reconstruction loss. The content-latent-code encoder is totally shared by two domains, in order to maintain the common content of two domains. It is composed of low-level layers $E_L$ and high-level layers $E_H^c$. The style-latent-code encoders are partly-shared by two domains, so that they capture different aspects of images from two domains. The shared part is also the low-level layers of the content-latent-code encoder, while the high-level layers $E_{A,H}^s$, $E_{B,H}^s$ are private. Three discriminators $D^c$, $D_A^s$ and $D_B^s$ are designed to tell whether the input is drawn from the corresponding prior distribution. Two generators $G_A$ and $G_B$ are used to generate images of domain $A$ and $B$ respectively.

\subsection{Training Method}

To train the framework, we mainly have two targets: one is that the output of encoders follows the corresponding prior distribution, and the other is that the generators can reconstruct the corresponding image from a given latent code vector.

To achieve the first target, we use an adversarial loss to impose prior distributions on the latent code vectors. Losses for encoders and discriminators are as follows:
\begin{equation}\begin{split}
  L_{adv}^s=&\alpha_1E_{\boldsymbol{x}_A \sim P_A(\boldsymbol{x}_A)}[\log(1-f_A^{d,s}(f_A^{e,s}(\boldsymbol{x}_A)))]\\+&\alpha_2E_{\boldsymbol{x}_B \sim P_B(\boldsymbol{x}_B)}[\log(1-f_B^{d,s}(f_B^{e,s}(\boldsymbol{x}_B)))]\\+&\alpha_3E_{\boldsymbol{z}_A^s \sim Q_A^s(\boldsymbol{z}_A^s)}[\log(f_A^{d,s}(\boldsymbol{z}_A^s))]\\+&\alpha_4E_{\boldsymbol{z}_B^s \sim Q_B^s(\boldsymbol{z}_B^s)}[\log(f_B^{d,s}(\boldsymbol{z}_B^s))],
\end{split}\end{equation}
\begin{equation}\begin{split}
  L_{adv}^c=&\beta_1E_{\boldsymbol{x}_A \sim P_A(\boldsymbol{x}_A)}[\log(1-f^{d,c}(f^{e,c}(\boldsymbol{x}_A)))]\\+&\beta_2E_{\boldsymbol{x}_B \sim P_B(\boldsymbol{x}_B)}[\log(1-f^{d,c}(f^{e,c}(\boldsymbol{x}_B)))]\\+&\beta_3E_{\boldsymbol{z}^c \sim Q^c(\boldsymbol{z}^c)}[\log(f^{d,c}(\boldsymbol{z}^c))],
\end{split}\end{equation}
where $f$ denotes the corresponding function represented by neural network, and the superscript $e, d, g, s, c$ refer to "encoder", "decoder", "generator", "style", "content", while subscript $A, B$ refer to domain $A$ and $B$ respectively. $\alpha_1$, $\alpha_2$, $\alpha_3$, $\alpha_4$ and $\beta_1$, $\beta_2$, $\beta_3$ are weights of those terms. These two losses represent how well the discriminators can tell the latent code drawn from the prior distributions and those gained from the encoders apart. The generators are aimed to minimize them, while the discriminators are aimed to maximize them. Therefore the framework is trained with the following minimax game:
\begin{equation}
  \min_{(f_A^{e,s},f_B^{e,s},f^{e,c})}\max_{(f_A^{d,s},f_B^{d,s},f^{d,c})}(L_{adv}^s+L_{adv}^c).
\end{equation}
With adversarial loss, we only need to be able to sample from the prior distribution in order to impose it on the latent code vector of the encoders. Therefore, choosing the prior distribution of the latent code vector is quit flexible.

As for the reconstruction loss and its optimization, we have the following formulas:
\begin{equation}\begin{split}
  L_{rec}=&\gamma_1E_{\boldsymbol{x}_A \sim P_A(\boldsymbol{x}_A)}[d(\boldsymbol{x}_A,f_A^{g}([f^{e,c}(\boldsymbol{x}_A),f_A^{e,s}(\boldsymbol{x}_A)]))]\\+&\gamma_2E_{\boldsymbol{x}_B \sim P_B(\boldsymbol{x}_B)}[d(\boldsymbol{x}_B,f_B^g([f^{e,c}(\boldsymbol{x}_B),f_B^{e,s}(\boldsymbol{x}_B)]))],
\end{split}\end{equation}
\begin{equation}
  \min_{f_A^g,f_B^g,f^{e,c},f_A^{e,s},f_B^{e,s}}L_{rec},
\end{equation}
where $d$ denotes the distance function of two images, and $[\boldsymbol{x}_1,\boldsymbol{x}_2]$ means the concatenation of $\boldsymbol{x}_1$ and $\boldsymbol{x}_2$. $\gamma_1$ and $\gamma_2$ are weights of the two terms. $L_{rec}$ measures the difference of the original image and the reconstructed image, and our framework is aimed to minimize it.

At this point, we have introduced how to train the framework with unlabeled data. To make use of the labeled data, we let the prior distribution of content-latent-code be the distribution of category, which means that the content-latent-code vector is a one-hot vector and its dimension is the number of categories. When the data is unlabeled, we use Eq. (13) to train the content-latent-code encoder. When the data is labeled, we use the following supervised loss to train it:
\begin{equation}\begin{split}
  L_{sup}=&\lambda_1E_{\boldsymbol{x}_A \sim P_A(\boldsymbol{x}_A)}[f_{CE}(f^{e,c}(\boldsymbol{x}_A),l_{\boldsymbol{x}_A})]\\+&\lambda_2E_{\boldsymbol{x}_B \sim P_B(\boldsymbol{x}_B)}[f_{CE}(f^{e,c}(\boldsymbol{x}_B),l_{\boldsymbol{x}_B})],
\end{split}\end{equation}
where $f_{CE}$ denotes the cross entropy function and $l_x$ is the label of the image $\boldsymbol{x}$. $L_{sup}$ measures the cross entropy of content-latent-code and the label of images from two domains, with weights $\lambda_1$ and $\lambda_2$.

To improve the result of cross domain transformation, we introduce the content-latent-code-consistency loss, which means that we want the translated image have the same content-latent-code as the input image. Let $\boldsymbol{x}_{A->B}$ denote the generated image of domain $B$ with image from domain $A$ as input, and $\boldsymbol{x}_{B->A}$ denote the inverse one. We mainly consider two  situations to handle: data of two domains is unlabeled or is labeled.

If data of two domains is unlabeled, we have the following formula of the content-latent-code-consistency loss:
\begin{equation}\begin{split}
  L_{cc}^{un}=&\eta_1E_{\boldsymbol{x}_A \sim P_A(\boldsymbol{x}_A)}[f_{CE}(f^{e,c}(\boldsymbol{x}_A),f^{e,c}(\boldsymbol{x}_{A->B}))]\\+&\eta_2E_{\boldsymbol{x}_B \sim P_B(\boldsymbol{x}_B)}[f_{CE}(f^{e,c}(\boldsymbol{x}_B),f^{e,c}(\boldsymbol{x}_{B->A}))].
\end{split}\end{equation}

If data of two domains is labeled, we have the following content-latent-code-consistency loss:
\begin{equation}\begin{split}
  L_{cc}^{su}=&\eta_1E_{\boldsymbol{x}_A \sim P_A(\boldsymbol{x}_A)}[f_{CE}(f^{e,c}(\boldsymbol{x}_{A->B}),l_{\boldsymbol{x}_A})]\\+&\eta_2E_{\boldsymbol{x}_B \sim P_B(\boldsymbol{x}_B)}[f_{CE}(f^{e,c}(\boldsymbol{x}_{B->A}),l_{\boldsymbol{x}_B})].
\end{split}\end{equation}

$L_{cc}^{un}$ and $L_{cc}^{su}$ measure the cross entropy of cross-domain generated images' content latent code and that of original images (labels when data is labeled) with weights $\eta_1$ and $\eta_2$ .We minimize this loss on the content-latent-code encoder and generators, in order to make the cross-domain generated image have the same content latent code with the original ones.

\subsection{Domain Adaptation}

The target of domain adaptation is to generalize the learned model of source domain to a target domain, where data of source domain is labeled while that of target domain is not labeled. Since CDAAE can be trained with both labeled and unlabeled data, there is an innate advantage for it to implement domain adaptation. Therefore, we design an algorithm based on CDAAE to perform domain adaptation.

\begin{algorithm}
  \caption{Basic Domain Adaptation}
  \label{bda}
  \begin{algorithmic}
    \STATE $S$ is labeled data of source domain, $T$ is unlabeled data of target domain
    \STATE $t \in [0,1]$ is the threshold of prediction probability
    \FOR{$i=1 \cdots pretrain\_steps$}
    \STATE $\min L_{sup}$ for $[\boldsymbol{x}_A, l_A] \in S$
    \ENDFOR
    \FOR{$i=1 \cdots train\_epoches$}
    \STATE $T'$=\{ $[\boldsymbol{x}, l]$ | $\boldsymbol{x} \in T$ and probability of $\boldsymbol{x}$'s label being $l$ > t \}
    \STATE $\min(L_{adv}^s+L_{adv}^c+L_{rec}+L_{cc}^{suun})$ for $[\boldsymbol{x}_A, l_A] \in S, \boldsymbol{x}_B \in T$
    \STATE $\min L_{sup}$ for $[\boldsymbol{x}_A, l_A] \in S, [\boldsymbol{x}_B, l_B] \in T'$
    \ENDFOR
  \end{algorithmic}
\end{algorithm}

\begin{algorithm}
  \caption{Domain Adaptation with Boosted Threshold}
  \label{da-boosted}
  \begin{algorithmic}
    \STATE $S$ is labeled data of source domain, $T$ is unlabeled data of target domain
    \STATE $t\_init \in [0,1]$ is the initial threshold of prediction probability
    \STATE $t=t\_init$ is the threshold of prediction 
    \STATE $w$ is a scale factor of the boosting rate of threshold
    \FOR{$i=1 \cdots pretrain\_steps$}
    \STATE $\min L_{sup}$ for $[\boldsymbol{x}_A, l_A] \in S$
    \ENDFOR
    \FOR{$i=1 \cdots train\_epoches$}
    \STATE $T'$=\{ $[\boldsymbol{x}, l]$ | $\boldsymbol{x} \in T$ and probability of $\boldsymbol{x}$'s label being $l$ > t \}
    \STATE $\min(L_{adv}^s+L_{adv}^c+L_{rec}+L_{cc}^{suun})$ for $[\boldsymbol{x}_A, l_A] \in S, \boldsymbol{x}_B \in T$
    \STATE $\min L_{sup}$ for $[\boldsymbol{x}_A, l_A] \in S, [\boldsymbol{x}_B, l_B] \in T'$
    \STATE $t=t\_init+(1-t\_init)(1-exp(-i \div w))$
    \ENDFOR
  \end{algorithmic}
\end{algorithm}

\subsubsection{Method}

As the content-latent-code encoder in CDAAE can be implemented as a classifier when its prior distribution is the distribution of category, we can train it as a classifier with labeled data from source domain and unlabeled data from target domain. Although the classifier trained with labeled data from source domain may not have a high accuracy on data from target domain, its prediction is still meaningful. Therefore we can use its prediction as a inexact label of data from target domain. So our domain adaptation algorithm has mainly two phases. In the first phase, we train the content-latent-code encoder as a classifier with labeled data from source domain, and then use its prediction on data from target domain as inexact label to form an inexactly labeled dataset of target domain. In the second phase, the whole framework is trained in a semi-supervised manner with labeled data of source domain, inexactly labeled data of target domain and unlabeled data of target domain. The algorithm is shown in Algorithm~\ref{bda}. To perform content-latent-code-consistency loss on data with only one domain labeled, we combine the first term of $L_{cc}^{su}$ and the second term of $L_{cc}^{un}$ to form $L_{cc}^{suun}$ as follows:
\begin{equation}\begin{split}
  L_{cc}^{suun}=&\eta_1E_{\boldsymbol{x}_A \sim P_A(\boldsymbol{x}_A)}[f_{CE}(f^{e,c}(\boldsymbol{x}_{A->B}),l_{\boldsymbol{x}_A})]\\+&\eta_2E_{\boldsymbol{x}_B \sim P_B(\boldsymbol{x}_B)}[f_{CE}(f^{e,c}(\boldsymbol{x}_B),f^{e,c}(\boldsymbol{x}_{B->A}))].
\end{split}\end{equation}

Since the ability of the classifier on data from target domain improves in the training process, we can raise the threshold to form a more exact inexactly-labeled data of target domain, in order to supply better directions for the training of classifier. Therefore, in farther, we propose the domain adaptation algorithm with boosted threshold, which is shown in Algorithm~\ref{da-boosted}.

\section{Experiments}

The Cross-Domain Adversarial Auto-Encoder (CDAAE) is mainly evaluated in two aspects. Firstly, we evaluate its ability to learn inference, generation and transformation of data from two domains. Secondly, we evaluate the domain adaptation algorithm based on CDAAE on benchmark datasets, which shows a considerable improvement over the previous state-of-the-art work.

\begin{figure*}[!t]
  \centering
  \includegraphics{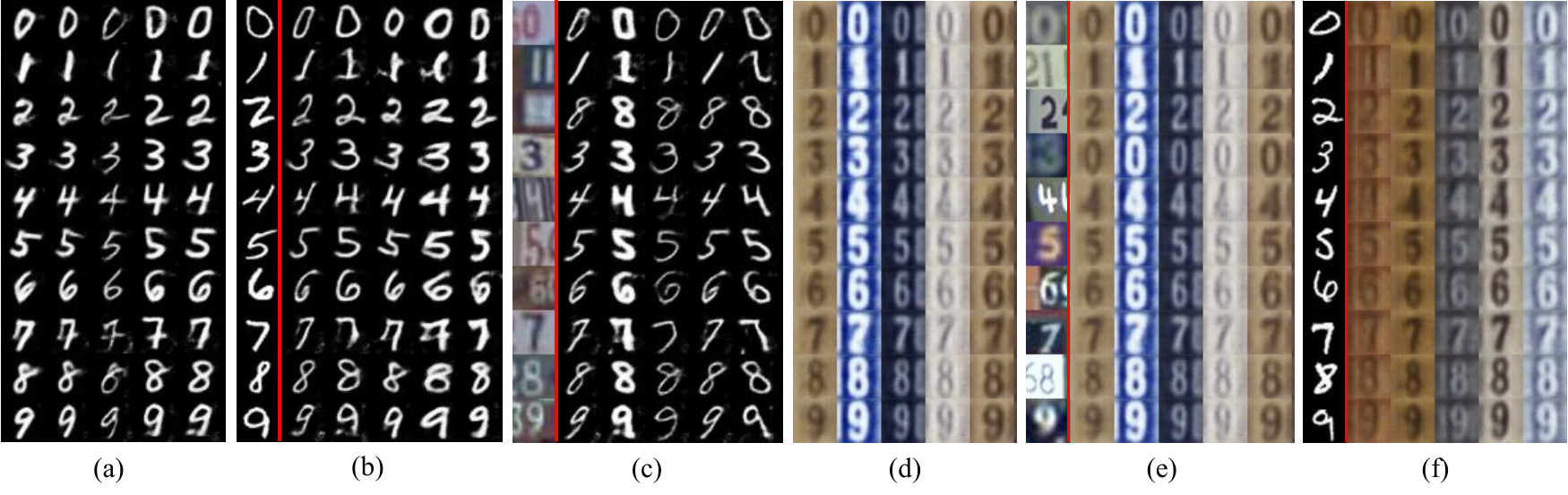}
  \caption{Above are the images generated by the model learned in supervised way. Each column of (a) and (d) represents generated images with latent code drawn from prior distribution, where the style latent code is fixed to a particular value but the content latent code is systematically explored. (b), (e) and (c), (f) demonstrate generated images with content latent code drawn from input images, which is shown in the left most column, separated by a red line, and style latent code drawn from prior distribution.}
  \label{supervised}
\end{figure*}

\begin{figure*}[!t]
  \centering
  \includegraphics{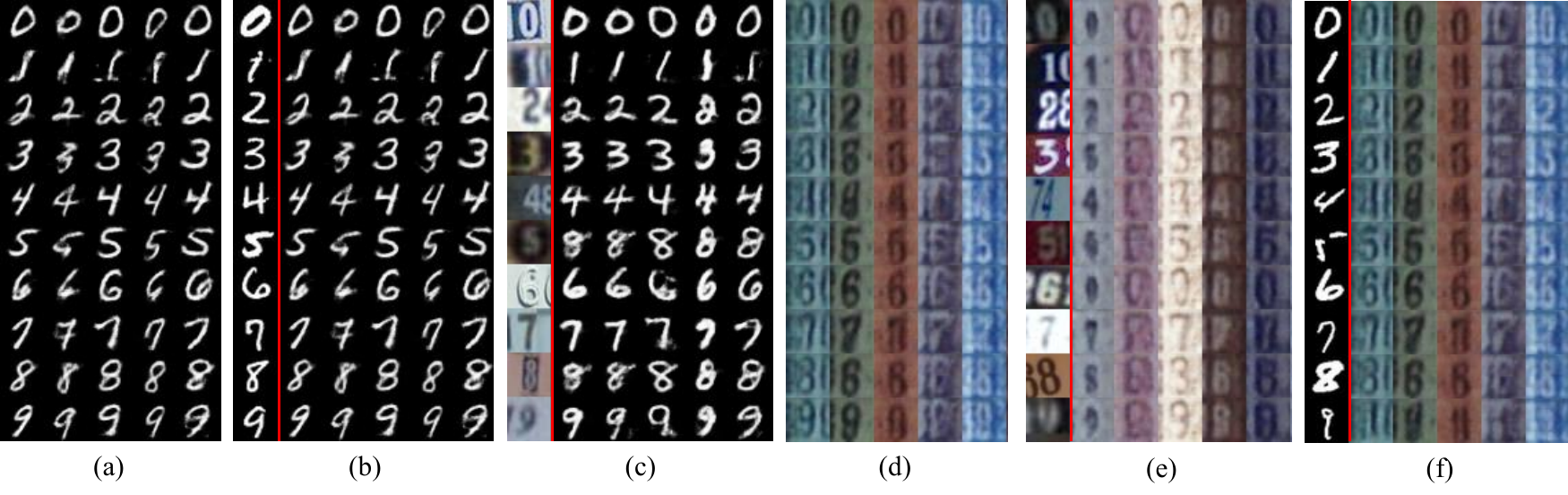}
  \caption{Above are the images generated by the model learned in semi-supervised way, with the same setting as the supervised-learned one when generating these samples. The images are demonstrated in the same configuration of Figure~\ref{supervised}.}
  \label{semi}
\end{figure*}

As for the first aspect, we evaluate the CDAAE in two application domains: digit and face images. For digit images, we use the Street View House Number (SVHN) dataset~\cite{svhn} and MNIST dataset~\cite{mnist} as domain $A$ and $B$ respectively. For face images, we use the CASIA NIR-VIS 2.0 Face Database~\cite{casia} with VIS images as domain A and NIR images as domain B. We show that CDAAE can learn an approximate inference of data from two domains, and is able to generate visually appealing samples in image generation and transformation. Besides we also demonstrate that CDAAE can generate various samples with a certain input, and the latent code representation is well compressed from the original image.

For domain adaptation, we evaluate our method on object classification datasets used in previous work, including SVHN~\cite{svhn}, MNIST~\cite{mnist} and USPS~\cite{usps}. We show that our method makes a considerable progress over the previous work.

\subsection{Digits: SVHN-MNIST}
\label{subsection:digits}

For experiments on digit images, we employ the extra training split of SVHN which contains 531,131 images and the training split of MNIST which contains 60,000 images as training set. The evaluation is done on the test split of SVHN and MNIST which contains 26,032 images and 10,000 images respectively. To make the MNIST images have the same shape as SVHN, we resize them to $32\times32$ and replicate the grayscale image three times.

\begin{figure*}[!t]
  \centering
  \includegraphics{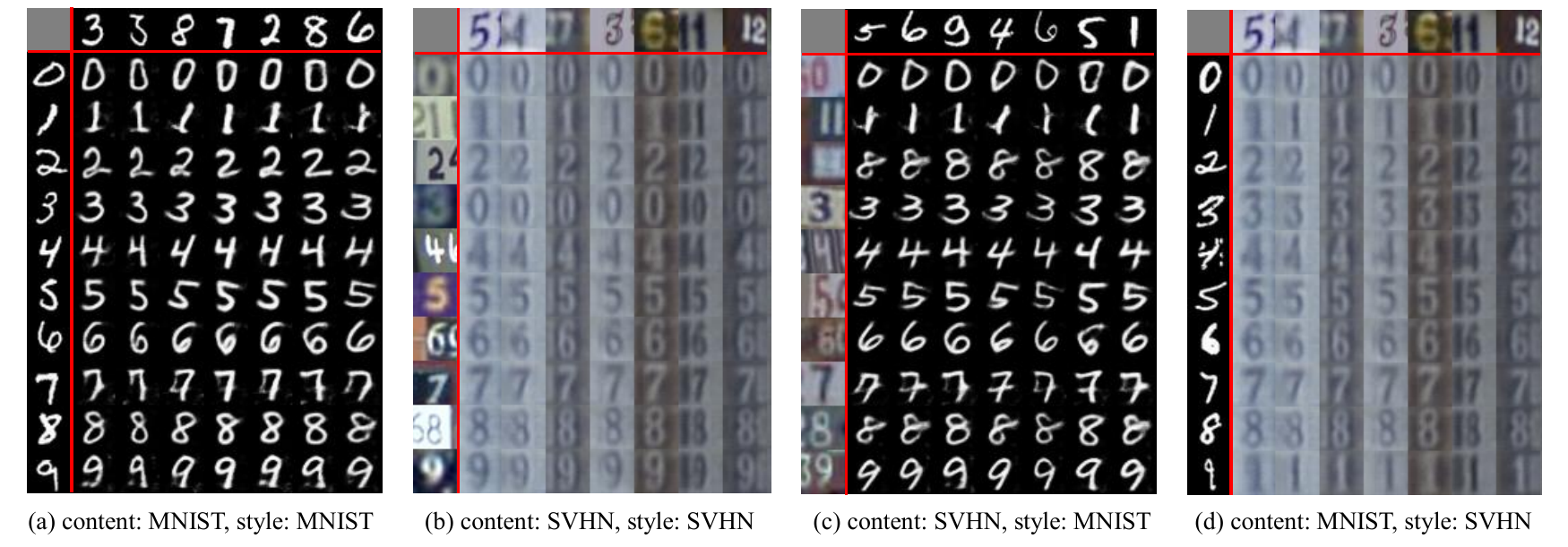}
  \caption{Images generated with content and style from two different images. Separated by red lines, the left most columns are the images supplying content latent code, while the top most rows are the images supplying style latent code. As can be seen, the images in the middle can keep the content with different styles learned from other images.}
  \label{style}
\end{figure*}

To make use of labeled data, we impose category distribution (10-D one-hot vector) on the content latent code and 8-D Gaussian on the style latent code for both SVHN and MNIST respectively.  Inspired by DTN, the encoders are implemented with four convolutional layers, followed by batch normalization and ReLU non-linearity. Layers of the content-latent-code encoder consists of 64, 128, 256, 128 filters respectively, and the last layer is followed by a fully connected layer with softmax to generate one-hot vectors. As for the style-latent-code encoders, they are built on the activations of the content-latent-code encoder's second layer, with three convolutional layers followed by batch normalization and ReLU, with the last layer not followed by ReLU to generate multi-dimensional Gaussian. The generators both contain four transposed convolutional layers with batch normalization and ReLU except the last layer which is activated by hyperbolic tangent. Besides, the three discriminators are implemented with four fully connected layers activated by ReLU. More details can be found in our code.

\subsubsection{Supervised Learning}

In supervised learning, training with $L_{sup}$ of Eq. (16) is enough for the content-latent-code encoder, so we set $\beta_1=\beta_2=0$. As for the other hyper parameters, we set $\gamma_1=2.0$, $\gamma_2=0.15$, $\lambda_1=5$, $\lambda_2=0.5$, $\eta_1=\eta_2=0.3$ and the remaining parameters are set to 1. As all the data is labeled, the content-latent-code-consistency loss for unlabeled data $L_{cc}^{un}$ is not used in this experiment. The experiment shows that our framework can generate visually appealing images from latent code drawn from prior distribution, and can generate various images with only one certain input in task of cross-domain transformation. Results are shown in Figure~\ref{supervised}.

\begin{table*}[!t]
  \centering
  \caption{Accuracy of classifiers on generated images (P denotes prior distribution, M denotes MNIST, S denotes SVHN)}
  \label{acc-trans}
  \begin{tabular}{c c c c c c c}
    \hline
    Method & P2M & M2M & S2M & P2S & S2S & M2S \\
    \hline
    Supervised CDAAE & \textbf{97.75\%} & 94.27\% & 83.21\% & \textbf{98.43\%} & \textbf{83.85\%} & \textbf{94.91\%} \\
    Semi-Supervised CDAAE & 97.57\% & \textbf{95.73\%} & 66.95\% & 78.24\% & 55.50\% & 78.67\% \\
    DTN~\cite{dtn} & -- & -- & \textbf{90.66\%} & -- & -- & -- \\
    \hline
  \end{tabular}
\end{table*}

\subsubsection{Semi-Supervised Learning}

In semi-supervised learning, we randomly sample 100 images for each category of each domain from the training data to form a labeled training dataset of totally 2000 images, and the remaining training data forms the unlabeled training dataset. The content-latent-code encoder is trained with both $L_{sup}$ of Eq. (16) for the labeled data and $L_{adv}^c$ of Eq. (12) for the unlabeled data. As for the content-latent-code-consistency loss, we use $L_{cc}^{su}$ of Eq. (18) for the labeled data and $L_{cc}^{un}$ for the unlabeled data. We did the same experiments on the learnt model as the supervised one. Results are shown in Figure~\ref{semi}. Though the model is trained with few labeled data, We can see that the generated images are visually appealing. The content latent codes has a right correspondence with the image, and the style latent code is meaningful.

\subsubsection{Evaluation of Generated Images}

To support quantitative evaluation, we trained a classifier for each domain with the training data, using the same architecture as the content-latent-code encoder. Different from the encoder, the two classifiers are trained separately for the two domains. The classifier for SVHN achieves an accuracy of 93.89\% on the test set, and the one for MNIST achieves an accuracy of 99.17\% on the test set. Therefore they are trustworthy as evaluation metrics. Samples generated with latent code vectors drawn from prior distribution are labeled with the corresponding content latent code, and we generate 1000 samples for each category for evaluation. We also use the test set of SVHN and MNIST as input to generate samples, whose content latent code is drawn from the input sample while style latent code is drawn from prior distribution. These samples is labeled with the label of the corresponding input samples. We run the classifiers on images generated by our framework, and the result is reported in Table~\ref{acc-trans}. From Table~\ref{acc-trans} though our accuracy is not as high as DTN, our framework has mainly two advantages over DTN. One is that our framework can do transformation bidirectionally, while DTN is specially trained for one way transformation. Another is that our framework can generate various samples with a certain input, while DTN can only generate one fixed sample.

\subsubsection{Analysis from an Information Theory View}

From view of information theory, the encoding-decoding process in image inference and generation corresponds to compression of image and decompression of compressed image. When the recovered image is well enough, the bigger data compression ratio, the better. In our framework, the content latent code is a 10-D one-hot vector, with a storage cost of 10 bits. The style latent code is a 8-D vector for both SVHN and MNIST, with a storage cost of $8\times32$ bits (the dtype is float32). With the original image of SVHN and MNIST having storage cost of $32\times32\times3\times8$ and $32\times32\times8$, our framework has a compression ratio of 92.39 and 30.80 respectively, while that of DTN~\cite{dtn}, whose latent code is 128-D float32 vector, is 6 and 2.

\begin{figure*}[t]
  \centering
  \includegraphics{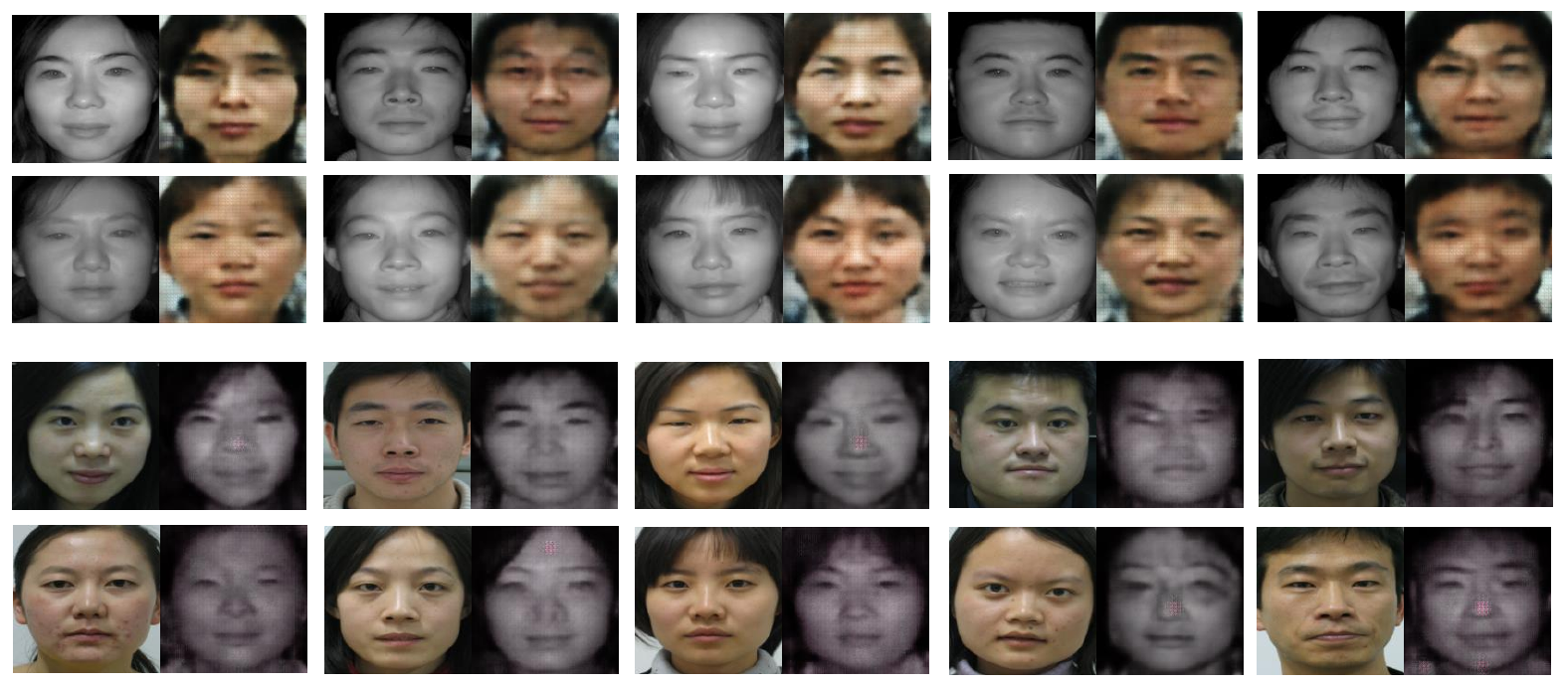}
  \caption{Results of transformation between face images of VIS and NIR. The top two rows are transformation from NIR to VIS, while the bottom two rows are transformation from VIS to NIR. For each pair, the left one is the input image, while the right one is the generated image.}
  \label{vis-nir}
\end{figure*}

\subsubsection{Style Transformation}

One of the main advantages of our framework is that we use separate style latent code, which is learned in a totally unsupervised manner, to represent the information except content. Therefore we can generate various images of one domain with only one certain input image of the other domain. Besides, we can also complete style transformation by concatenating content latent code and style latent code of different images and feeding the result into the corresponding generator. We evaluate this by combine content and style from different images within domain and cross domain, which leads to four scenarios, as shown in Figure~\ref{style}. The visually appealing results illustrate that the style representation learned by our framework is convincing and meaningful.

\begin{table*}[!t]
  \centering
  \caption{Evaluation of domain adaptation: accuracy on test split of target domain}
  \label{acc-adapt}
  \begin{tabular}{c c c c c}
    \hline
    Method & SVHN to MNIST & USPS to MNIST & MNIST to USPS\\
    \hline
    CORAL~\cite{coral} & -- & -- & 81.7\%~\cite{pixelDA} \\
    MMD~\cite{MMD-2,MMD-1} & -- & -- & 81.1\%~\cite{pixelDA} \\
    DANN~\cite{dann} & 73.85\% & -- & 85.1\%~\cite{pixelDA} & \\
    DSN~\cite{dsn} & 82.7\% & -- & 91.3\%~\cite{pixelDA} \\
    CoGAN~\cite{CoGAN} & -- & 89.1\% & 91.2\% \\
    PixelDA~\cite{pixelDA} & -- & -- & 95.9\% \\
    DTN~\cite{dtn} & 84.44\% & -- & -- \\
    SA~\cite{domainAdapt-subspace} & 59.32\% & -- & -- \\
    UNIT~\cite{unI2I} & 90.53\% & 93.58\% & \textbf{95.97\%} \\
    CDAAE(basic) & 96.67\% & \textbf{96.37\%} & 95.91\% \\
    CDAAE(boosted) & \textbf{98.28\%} & 95.89\% & 95.86\% \\
    \hline
  \end{tabular}
\end{table*}

\subsection{Faces: CASIA VIS-NIR}

The CASIA VIS-NIR dataset~\cite{casia} is a face image dataset with two domains: the near infrared (NIR) and visible light (VIS) images. There are totally 725 classes, and we selected those classes with more than five images for both VIS and NIR for our experiment, which consists of 582 classes. We selected one or two images of each class for both domains as the test data, and the remaining forms the training data.

The architecture is similar with that of digits, but we expanded the encoders to six layers and the generators to seven layers. We use a 512-D Gaussian as the content latent code and a 64-D Gaussian as the style latent code. The supervised information is used through an additional fully connected layer followed by softmax on top of content latent code. The result of cross-domain transformation is shown in Figure~\ref{vis-nir}. As can be seen, our framework can achieve cross-domain transformation. Reasons of content not maintained so well may be that samples for each class is few.

\subsection{Domain Adaptation}

The implement of the framework here is same as the one in Section~\ref{subsection:digits}. We evaluate the proposed domain adaptation methods with the following scenarios: SVHN to MNIST, USPS to MNIST and MNIST to USPS. In all scenarios, the model is trained with the labeled training split of source domain and unlabeled training split of target domain. Models are evaluated with the accuracy on the test split of target domain. USPS has 7291 images for training and 2007 images for testing. We resize all images to $32\times32$. For MNIST, we resize it directly, while for USPS, we firstly center it in a $22\times22$ image, and then resize the $22\times22$ image to $32\times32$. Grayscale images are replicated three times. In all three scenarios, we use 16-D Gaussian for style latent code of source domain, and 8-D Gaussian for that of target domain. The threshold of prediction probability in the basic algorithm (or initial threshold in the boosted one) is set to 0.85, and the scale factor of the boosting rate in boosted algorithm is set to 10,000. Other hyper parameters are the same as Section~\ref{subsection:digits}. Results are reported in Table~\ref{acc-adapt}. We can see that in scenario of MNIST to USPS, we achieve comparative accuracy to the state-of-the-art work. However in more difficult scenarios: SVHN to MNIST and USPS to MNIST, we make a considerable progress over the previous state-of-the-art work.

\section{Conclusion and Future work}

In this paper, we proposed a novel framework named Cross-Domain Adversarial Auto-Encoder (CDAAE), which can complete inference, generation and transformation of cross domain images. We made the assumption "shared content-latent-code space while separate style-latent-code space", in order to disentangle the representation of content and style. This allows us to generate various samples with just one certain input, which can be applied to cross-domain transformation, style transfer and data augmentation. Besides, the proposed framework can be trained with both labeled and unlabeled data, so a domain adaptation algorithm is designed, which achieves state-of-the-art accuracy on benchmark datasets.

Since experiments are mainly performed on image datasets, more application domains, such as audio and text, can be explored. As only the reconstruction loss and content-latent-code-consistency loss are used to train the generator, it's worth to try adding adversarial loss on the generator to improve the generated images. Besides, as the prior distribution is quite flexible in our framework, more other kinds of prior distributions can be explored.
%\end{document}  % This is where a 'short' article might terminate

\bibliographystyle{ACM-Reference-Format}
\bibliography{sample-bibliography}

\end{document}